\pdfoutput=1

\documentclass[11pt]{article}
\usepackage{cjhebrew}
\usepackage{balance} %

\usepackage[]{acl}

\usepackage{times}
\usepackage{latexsym}

\usepackage[T1]{fontenc}

\usepackage[utf8]{inputenc}

\usepackage{microtype}

\usepackage{graphicx}
\graphicspath{ {./plots/} }
\usepackage{booktabs}
\usepackage{multicol}
\usepackage[compact]{titlesec}
\usepackage{url}

\title{A Dataset for Metaphor Detection in Early Medieval Hebrew Poetry}

\author{Michael Toker\textsuperscript{1} \hspace{1em} Oren Mishali\textsuperscript{1} \hspace{1em} Ophir Münz-Manor\textsuperscript{2} \\ {\bf Benny Kimelfeld}\textsuperscript{1} \hspace{1em} {\bf Yonatan Belinkov}\textsuperscript{1} \\
\textsuperscript{1} Technion -- Israel Institute of Technology \hspace{2em} 
\textsuperscript{2} The Open University of Israel  \\
        \texttt{\{tok,omishali,bennyk,belinkov\}@cs.technion.ac.il} \\
        \texttt{ophirmm@openu.ac.il}}

\begin{document}
\maketitle
\begin{abstract}
There is a large volume of late antique and medieval Hebrew texts. They represent a crucial linguistic and cultural bridge between Biblical and modern Hebrew. 
Poetry is prominent in these texts and one of its main characteristics is the frequent use of metaphor. Distinguishing figurative and literal language use is a major task for scholars of the Humanities, especially in the fields of literature, linguistics, and hermeneutics.
This paper presents a new, challenging dataset of late antique and medieval Hebrew poetry with expert annotations of metaphor, as well as some baseline results, which we hope will facilitate further research in this area.\footnote{Code, data and demo are available in project website \href{https://tokeron.github.io/metaphor/}{tokeron.github.io/metaphor}.} %
\end{abstract}

\section{Introduction}
The Hebrew language has a long and rich history, from Biblical Hebrew, through Rabbinic-Medieval Hebrew, to modern Hebrew.
In this work, we present a corpus consisting of Hebrew liturgical poetry from the fifth to eighth centuries CE, also known as Piyyut (from Greek \emph{poietes}, to create, versify; plural: Piyyutim).
The Piyyutim in the corpus were reconstructed throughout most of the twentieth century by various scholars from manuscripts of the Cairo Genizah, a medieval repository of Jewish texts  \cite{van2008hebrew, theLostArchive}.
Since poetry was a prominent genre in late antique and medieval Hebrew literature, the corpus is rich in figures of speech like similes and metaphors. 

Active research in this area is conducted by scholars in the Humanities, especially Digital Humanities, who wish to understand not only the literal meaning of a text but also its figurative meaning \cite{munz2011figurative}. At present, texts are annotated manually, a time-consuming and labor-intensive process. 
Scholars of Hebrew literature and Hebrew linguists would thus benefit greatly from a tool that automatically detects figurative language in these texts. %
Furthermore, such tools could be used by non-specialists who want to better understand these texts by highlighting figurative language. Since the literary and linguistic tradition of Piyyut runs throughout the Middle Ages, working on the early strata of this tradition would enable us to extend the impact of metaphor detection also to later periods and other genres. 
Yet, to the best of our knowledge, there are no previous studies that deal with this task, in either modern or pre-modern Hebrew.

To fill this gap, the main contribution of this work is a medieval Hebrew dataset of Hebrew liturgical poetry with metaphor annotations.
The dataset consists of two units of ancient Piyyut, with 309 poems and 73,179 words, with expert annotations for metaphorical expressions. 
Despite its relatively small size, the corpus contains 15\% of the digitized Piyyutim and is the only metaphor-annotated corpus available in Hebrew.

We develop and evaluate several transformer-based models for detecting metaphors in the dataset, based on two pre-trained Hebrew language models:  AlephBERT, which was pre-trained on modern Hebrew  \cite{seker2021alephbert}, and BEREL,  pre-trained on ancient Jewish texts that are closer in style to the Piyyut texts  \cite{shmidman2022introducing}. 
We substantially improve na\"ive baselines, with our best model achieving F1 scores of 48.7 and 49.4 on the two corpora. Considering the difficulty of the task, attested through an inter-annotator agreement study we conducted,  
we find the results encouraging while leaving ample room for improvements.

\section{Background}
\subsection{Literary and Linguistic Background}

Jewish liturgy took shape in the Near East in the first centuries of the Common Era and by the end of the 3rd century began to take on fixed forms. In the late 4th century, poets began to embellish liturgical prose, infusing religious meaning with poetic beauty. By the 7th century, Piyyut (Jewish liturgical poetry) became an integral medium of religious discourse
and Payytanim (liturgical poets) evolved into prominent cultural figures \cite{lieber2010yannai}.

The study of Piyyut is relatively young and rather 
small in scale, since most of the Payytanic texts from this period were discovered towards the end of the 19th century in the Cairo Genizah. Throughout most of the twentieth-century scholars of Piyyut focused on literary and linguistic investigations of the texts \cite{van2008hebrew}.
In essence, the Payytanic language constitutes a separate stratum in the history of the Hebrew language although it is much closer to biblical Hebrew than to contemporaneous Rabbinic Hebrew. Importantly, there are significant differences between Piyyut and modern Hebrew, at syntactic and lexical levels. 

In summary, metaphor plays an important role in the literary fabric of Piyyut and at later stages, most notably in the Islamic East, metaphorical expressions become increasingly central and innovative. The study of figurative language in Piyyut and more broadly in medieval Hebrew literature remains a major task. Computational tools would greatly help advancing this area \cite{munz2011figurative}.

\subsection{Hebrew NLP} \label{sec:hebrewnlp}
Hebrew is a low-resourced morphologically-rich language with few labeled datasets, which are typically in modern Hebrew  \cite{keren2021parashoot,litvak2022offensive}. 
Notable unlabeled Hebrew corpora are the Ben-Yehuda project \cite{BenYehudaProject}, a heterogeneous collection of medieval and modern Hebrew literature;
and the Sefaria \cite{sefaria} 
and Dicta Library \cite{Dicta}
collections of ancient Jewish texts.

Several  Hebrew language models have been released, most of them trained on limited data compared to English language models \cite[e.g., HeBERT;][]{chriqui2021hebert}.
A prominent model is AlephBERT \cite{seker2021alephbert},  which was trained on 1.9 billion words of modern Hebrew. Fine-tuning it led to high performance on multiple sequence labeling tasks. %
A more recent model is BEREL \cite{shmidman2022introducing}. It was pre-trained on Rabbinic Hebrew texts from Sefaria and the Dicta Library, which are more similar to Piyyut  than  modern Hebrew. 
BEREL's training set is an order of magnitude smaller than  AlephBERT's (220 million compared to 1.9 billion words). %

\subsection{Metaphor Detection}
Metaphor detection is the task of identifying metaphorical expressions in natural language. In this section, we review some of the existing computational approaches to metaphor detection.

One of the earliest computational approaches to metaphor detection is based on the notion of Selectional Preference Violation (SPV) \cite{wilks1975preferential}. SPV occurs when a word or a phrase differs from its typical or expected domain of usage, indicating a possible non-literal meaning. Based on this idea, \citet{fass-1991-met} developed met*, one of the first systems to automatically identify metaphorical expressions in text, using hand-coded knowledge and SPV as indicators of non-literalness. Later, \citet{mason-2004-cormet} presented CorMet, the first system to automatically discover source–target domain mappings for metaphors, by detecting variations in domain-SPV from Web texts.

Another computational approach to metaphor detection is based on the use of different linguistic features. One example is the notion of abstractness and concreteness. Abstractness and concreteness are semantic properties of words or concepts that reflect their degree of perception or imagination. For example, the word ``love'' is more abstract than the word ``rose'', because the former is less perceptible or imaginable than the latter. Based on this idea, \citet{turney-etal-2011-literal} proposed a method to detect metaphorical usage by measuring abstractness and concreteness. Other feature-based methods include semantic supersenses \citep{tsvetkov-etal-2013-cross} and imageability \citep{10.1007/978-3-642-37210-0_12}

However, both SPV-based and feature-based approaches have some limitations. One of the main limitations is that they fail to generalize well to rare or novel metaphorical uses, because they rely on predefined or precomputed knowledge or features. To overcome this limitation, more recent approaches have explored the use of learned representations to detect metaphors. \citet{shutova-etal-2016-black} proposed a method to detect metaphors by using a set of arithmetic operations on learned word representations. For details, refer to \citet{Metaphor} and \citet{shutova-etal-2013-statistical}.

More recently, some studies have focused on metaphor detection with pre-trained English transformers \citep{vaswani2017attention}. Transformers are a type of neural network that can encode and decode sequences of words or symbols using attention mechanisms, which allow them to focus on the most relevant parts of the input or output. Pre-trained transformers are transformers that have been trained on large amounts of text data, such as Wikipedia or news articles, and can be fine-tuned or adapted to specific tasks or domains. 

\citet{gong-etal-2020-illinimet} use RoBERTa, a pre-trained transformer with rich linguistic information from external resources such as
WordNet, to train a feed forward layer to identify whether a given word is a metaphor. Another work \citep{liu-etal-2020-metaphor} uses both BERT and XLNet language models to create contextualized embeddings and a bidirectional LSTM for the same task. 

\citet{su2020deepmet} use augmented 
 BERT \cite{devlin2018bert} with local representations of candidate words and linguistic features such as part of speech. 
\citet{choi2021melbert} utilize the gap between the representation of a word in context and its absence, as well as the gap between the metaphor word and its neighbors.

We are not aware of any work on automatic metaphor detection in Hebrew in general and in pre-modern Hebrew specifically.

\begin{table*}[t]
    \centering
    \begin{tabular}{rll}
    \toprule
    Hebrew Source & Literal Translation & Meaning \\ \midrule 
    \<bgzrwt> \underline{\<.tb`nw>} & We \textbf{drowned} in decrees & There are too many decrees\\ 
    \underline{\<bm/sl.h ydnw>} \<`.sbwn>  & Irritation is \textbf{in our hands} & We are sad at work \\
    \<'.hqwr> \underline{\<qrbyym klywt>} \<'.hps> & I'll explore \textbf{kidney guts} & Investigate the true intentions\\
    \underline{\<wyb`r nr>} \<n/smh> \underline{\<h.syt>}  & \textbf{Ignite} a soul, \textbf{fire a candle} & Activate a soul \\ 
    \underline{\<`/sw pry>} \<l'> & Did not \textbf{bear fruit} & Did no good deeds \\ 
    \bottomrule
    \end{tabular}
    \caption{Examples from our dataset, with metaphorical expressions in underline/bold.}
    \label{tab:examples}
\end{table*}

\section{The Dataset}
\subsection{Construction and Annotation} \label{sec:construction}
The dataset consists of two separate corpora of Piyyut: (1) 172 poems by various poets (all anonymous except for one, Yosei ben Yosei) that were composed during the 5th century CE in the Galilee. This is the earliest corpus of Piyyut and it represents the formative phase of this poetic tradition, referred to here as Pre-Classical Piyyut. 
With an average of 1,213 words for a poem, and 1.64 words for metaphor phrases, the text length varies between 99 and 20,735 words.
(2) 137 poems by Pinchas Ha-Cohen (the Priest), who  lived in the first half of the 8th century CE in Tiberias, and is regarded as the last major poet of the classic payytanic tradition \cite{elizur2004liturgical}.
Text length ranges from 38 words to 9,683, with an average of 1,162 words. Metaphor phrase length averages 2.46 words.
Both corpora were recovered from medieval manuscripts that were unearthed towards the end of the 19th century in a medieval synagogue in Cairo. 

The entire corpus was manually analyzed and annotated by a Hebrew literature professor specializing in the study of Piyyutim, who studied the literary aspects of the corpus with a special emphasis on figurative language and metaphor in particular. It %
was digitized using the CATMA %
annotation tool \cite{catma}. Annotation has been done at the level of single words or multiword expressions, where the expert annotator highlighted a span of words corresponding to a single metaphor.  
Table~\ref{tab:examples} contains  examples of texts and metaphor annotations from the dataset. 

Since the identification of metaphor is to some extent interpretative, we asked another literary expert to annotate part of the corpora so we can calculate inter-annotator agreement and have a benchmark to evaluate the results of the models. (Annotator guidelines can be found in \ref{appendix:annotator_guidelines}.)
The second expert annotated 27.7\% of the first corpus (12,104 words) and 18.5\% of the second (5,454 words). The calculated  Cohen’s kappa scores  are 0.618 for Pre-Classical Piyyut and 0.628 for the  Pinchas corpus, which are similar to the 0.63 agreement reported by \citet{shutova-etal-2013-statistical} for English metaphor annotation. 
Although considered a ``substantial'' agreement, the score reflects non-negligible variations between the two annotators. A discussion about the inter-annotator disagreement including examples is given in appendix \ref{appendix:iaa}. It should be noted that while in some cases they are due to human error, in  more complex setups, variations are plausible and may be considered in modeling \cite{plank}.

\subsection{Statistics and Standard Splits} \label{sec:stats}
Descriptive statistics of the dataset are summarized in Table \ref{tab:dataset_size}. We note that 16.3\% and 21.3\% of the words are annotated as a metaphor in the Pre-Classical Piyyut and Pinchas corpora, respectively. A few texts have an unusual high percentage of metaphor usage (App.~\ref{sec:appendix-stats}). 

\begin{table}[h]
    \centering
    \begin{tabular}{l rr}
    \toprule
    & \multicolumn{1}{c}{Pre-Classical} & \multicolumn{1}{c}{Pinchas}  \\ 
    \midrule
    \# texts & $172$ & $137$\\ 
    avg text length & $1,213$ & $1,162$\\ 
    \# sentences & $6,836$ & $6,881$\\
    \% SM* & $38.28$ & $33.31$\\
    \# words & $43,697$ & $29,482$ \\ 
    \# metaphor & $7,123$ & $6,280$ \\
    \% metaphor & $16.3$ & $21.3$ \\
    \bottomrule
    \end{tabular}
    \caption{Overall statistics of the two corpora. SM* stands for sentences that contain at least one metaphor.}
    \label{tab:dataset_size}
\end{table}

To facilitate reproducible research with the dataset, we define standard splits to training, validation, and test sets (split 64/16/20\%, respectively). Table \ref{tab:metaphor_split} (App.~\ref{sec:appendix-stats}) has exact sizes. 
We randomly split by text, so each text is only found in one split. To ensure similar distributions across splits, we stratify by text length and metaphor ratio.

Of the words annotated as a metaphor in the test sets of Pre-Classical Piyyut and Pinchas, respectively, 55\% and 52\% do not appear as a metaphor in the corresponding training sets. 
Thus lexical memorization is not enough for this dataset.

\subsection{Limitations}
As aforementioned, metaphor detection involves human interpretation, making ambiguity common in both human and automatic metaphor detection. 

The Pre-Classical Piyyut corpus was reconstructed from an arbitrary collection. The poems we have are the only ones that survived from the 5th century and in most cases we cannot identify the poets. Therefore, the corpus is not homogeneous and its literary and linguistic aspects can differ considerably. Consequently, manual or automatic metaphor detection may become more challenging. The Pinchas corpus, in contrast, even if not complete because some poems may have been lost over time, represents the poetic works of one poet, hence it is much more homogeneous.

\section{Experimental Evaluation}

\subsection{Problem Formulation and Metrics}
We treat metaphor detection as a sequence labeling task, with each word labeled as metaphor (`M`) or non-metaphor (`O`). To represent multiword metaphors, we follow a BIO scheme where the first word is indicated with  ``B-M`, and the other words  with  `I-M`. Refer to App.~\ref{appendix:automatic_labeling} for more details.
Given the unbalanced nature of the dataset (Section \ref{sec:stats}), we focus on the F1 score, but also report precision, recall, and accuracy. 

\subsection{Naive Baselines}
Due to the novelty of this task, we report two na\"ive baselines. The majority baseline always assigns non-metaphor, obtaining around 80\%  accuracy, but its F1 score is zero. Another baseline is assigning the most frequent tag of the word in the training set for seen words, and a non-metaphor tag for unseen words. This baseline achieves a 24 F1 score. 
See Table \ref{tab:results-both} for F1 scores and other metrics in App.~\ref{sec:appendix-results}. In general, both corpora show similar trends. 

\subsection{Transformer-based models}
We experiment with two pre-trained Hebrew language models---AlephBERT and BEREL--- which we fine-tune on the metaphor detection task. Both models are encoder-only with 12 layers.
The two models differ in the pre-training data, as well as their tokenizers and vocabularies (50K items in AlephBERT, 128K items in BEREL). The results in this section are the average of five runs with different seeds. 
Details about the training and hyperparameters can in found in App.~\ref{sec:training}

To examine the effect of the tokenizers, we first trained randomly-initialized versions of the two models on metaphor detection, obtaining poor F1 results of about 30--34.

Next, we fine-tuned the pre-trained models, yielding substantial improvements: 40.8/42.2 F1 with AlephBERT on the two corpora, 43.7/46.5 with  BEREL.  We attribute the superior performance of BEREL both to its pre-training data being closer to the Piyyut language compared to AlephBERT's modern Hebrew pre-training data, and to its vocabulary size. It is especially impressive considering BEREL had ten times less training data.

The fact that BEREL outperforms AlephBERT despite being pre-trained on less data suggests that adaptation to the target genre is crucial. 
Following \citet{gururangan2020don}, 
we adapted AlephBERT to Piyyut by training it with  masked language modeling on texts more similar to Piyyut: first 
texts from Project Ben-Yehuda (approximately 2.7 million words.); then our Piyyut corpus (without labels). 
Finally, we fine-tuned the adapted model on metaphor detection. This step improved results by 1--2\%  (``adapted'' rows,  Table \ref{tab:results-both}). 

In view of the unbalanced data (metaphor words are only 16\% in Pre-Classical Piyyut and 21\% in Pinchas), we used a weighted cross-entropy (WCE) loss. By increasing the loss of the wrong prediction of the less frequent class (metaphor), we encourage the model to identify more words as a metaphor. This modification hurts precision and increases recall, resulting in an  increase in  F1 scores of 3--4 points (WCE rows in Table \ref{tab:results-both}; Tables \ref{tab:results-pre} and \ref{tab:results-pinchas} in App.~\ref{sec:appendix-results}). 
Fine-tuning BEREL with WCE provided the best results in terms of F1.
Furthermore, we examine the percentage of perfectly predicted words (correct prediction in all appearances). We find that 71\% of the words that appeared in the validation set were perfectly predicted.

\begin{table}
    \centering
    \begin{tabular}{l cc}
    \toprule 
    Model & Pre-Classical & Pinchas \\ 
    \midrule 
    Global majority 	&		 $0.0$ & 		 $0.0$ \\ 
    Most frequent tag 	&		 $24.2$ &		 $24.7$ \\
    \midrule 					
     BEREL rand  	&		 $30.7 \pm 2.1$ &    		 $34.4 \pm 2.3$ \\    
    AlephBERT rand  	&		 $31.6 \pm 2.2$ &		 $31.3 \pm 3.4$ \\
    \midrule 					
     BEREL 	&		 $43.7 \pm 0.6$&    		 $46.5 \pm 2.0$\\ 
    \hspace{5pt} + WCE 	&		 $\mathbf{48.7 \pm 1.4}$ &   		 $\mathbf{49.4 \pm 0.8}$ \\       
    AlephBERT 	&		 $40.8 \pm 2.0$ & 		 $42.2 \pm 1.2$ \\ 
    \hspace{5pt} + WCE 	&		 $45.9 \pm 0.7$ &		 $45.5 \pm 2.0$ \\
    \hspace{5pt} + adapted 	&		 $42.8 \pm 1.3$&		 $44.8 \pm 0.7$\\
    \hspace{5pt} + adapted+WCE 	&		  $47.2 \pm 0.9$  &		  $47.3 \pm 1.0$  \\
    \bottomrule 
    \end{tabular}
    \caption{Metaphor detection average F1 scores. Each experiment was repeated five times with different seeds. }
    \label{tab:results-both}
    \vspace{-5pt}
\end{table}

\subsection{Error Analysis}
\label{sec:error_analysis}
We examined how the best model (BEREL, trained with WCE) performs on words in the validation set (of the Pre-Classical corpus) that are not in the training set (``unseen'' words), compared to its performance on ``seen'' words  that exist in the training set. While the F1 score for seen words (54.6) is greater than unseen words (44.3), the latter score is still substantial, indicating that the model has learned to generalize to new words and metaphors.

We qualitatively analyzed the most common mistakes made by BEREL and AlephBERT models. Anecdotally, we found BEREL to better reflect metaphorical usage common in ancient texts, while AlephBERT tended to prefer literal meaning common in modern texts. As an example, consider the phrase \<n`ylt /s`r> that is used in Piyutim as a metaphor for the ``locked gate to the sky''. While this phrase is a common metaphor in ancient texts, its occurrence in modern Hebrew is notably diminished, predominantly confined to its literal interpretation. It appears that the BEREL model, trained on ancient texts, outperforms the AlephBERT model in capturing the metaphorical nuances of the phrase. The latter, trained on modern Hebrew, is likely more attuned to its literal interpretation.

Although many of the model's errors can be attributed to its inaccuracy, some accrue due to the sentence's ambiguity. For example, the sentence \<yld lny.hw.h .h/sqtyw> (A child to the smell of his desires) is annotated as literal by one annotator, and as metaphorical by the model. According to the expert annotator, the child here is a non-metaphorical nickname for Isaac. "smell" here is a non-metaphorical term for the victim. 'desires' can be metaphorical in a certain context, so it is not a complete mistake to claim that the sentence is metaphorical. See App.\ \ref{appendix:mistakes} for more details.

\section{Conclusion} 
We presented a corpus of medieval Hebrew poetry with metaphor annotations. The corpus can serve literary scholars who wish to study figurative language use in this genre. We also evaluated basic approaches for automatic metaphor detection based on this corpus, emphasizing the importance of adapting models to this particular genre. Models such as these have some practical applications. By automatically detecting metaphors in Piyut texts, people can better understand these ancient texts. Furthermore, these tools may allow Experts to semi-automatically annotate more texts.
We hope to facilitate further research in this area, both in designing more sophisticated methods for metaphor detection in this challenging corpus and in improving the workflow of literary scholars interested in this body of texts.

\section*{Acknowledgements}
This work was supported by the Israeli Ministry of Innovation, Science \& Technology (grant No.\ 3-17993), the ISRAEL SCIENCE FOUNDATION (grant No.\ 448/20), and an Azrieli Foundation Early Career Faculty Fellowship.

\bibliography{custom}
\bibliographystyle{acl_natbib}

\clearpage 

\appendix

\section{Appendix}
\label{sec:appendix}

\subsection{Additional Statistics} \label{sec:appendix-stats}

\begin{figure}[h]
    \centering
    \resizebox{0.95\columnwidth}{!}{\includegraphics{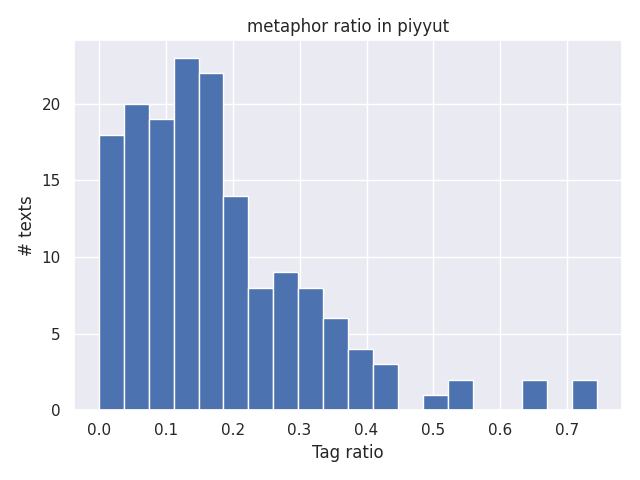}}
    \caption{Distribution of the metaphor ratio in the Pre-Classical Piyyut corpus.}
    \label{fig:metaphor_ratio_Piyyut}
\end{figure}

\begin{figure}[h]
    \centering
    \resizebox{0.95\columnwidth}{!}{\includegraphics{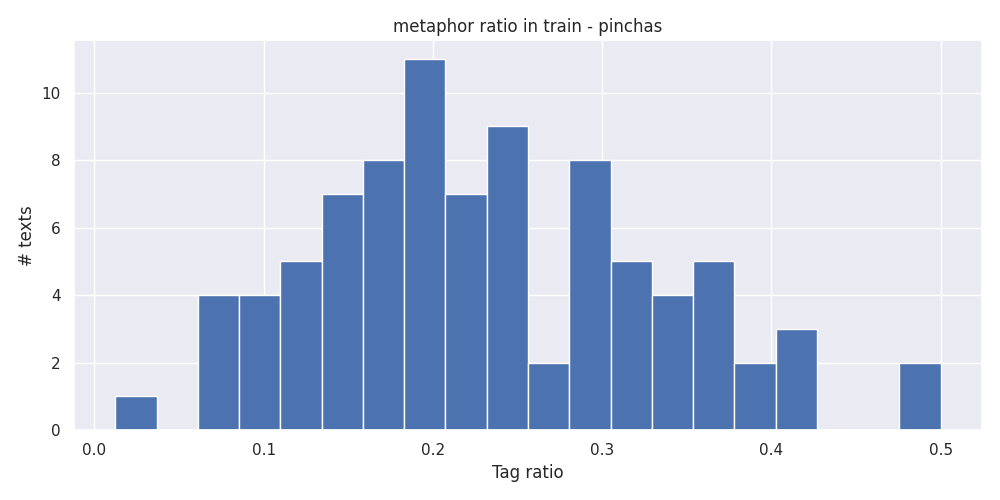}}
    \caption{Distribution of the metaphor ratio in the Pinchas corpus.}
    \label{fig:metaphor_ratio_Pinchas}
\end{figure}

Figures \ref{fig:metaphor_ratio_Piyyut} and \ref{fig:metaphor_ratio_Pinchas} show histograms of texts in the two corpora, binned by the ratio of metaphor words they contain. While a few texts contain a very high ratio of metaphor words, most texts have a small ratio. Table \ref{tab:metaphor_split} presents the division of the dataset into training, validation, and test splits.

\begin{table}[h]
    \centering
    \begin{tabular}{l ccc}
    \toprule 
     & Training & Validation & Test  \\ 
     \cmidrule(lr){2-4}
     & \multicolumn{3}{c}{Pre-Classical}  \\ 
     \cmidrule(lr){2-4}
    Metaphor & 4707	& 1070 & 1070  \\ 
    Non-Metaphor & 26485 & 26485 & 5103 \\
    Total & 31192 & 6322 & 6183  \\
    \midrule 
    & \multicolumn{3}{c}{Pinchas} \\ 
    \cmidrule(lr){2-4}
    Metaphor &  4105 & 867 & 1225 \\ 
    Non-Metaphor  & 15552 & 2932 & 4801\\
    Total & 19657 & 3799 & 6026 \\
    \bottomrule
    \end{tabular}
    \caption{Number of  tokens in each  split for each corpus.}
    \label{tab:metaphor_split}
\end{table}

\subsection{Intended Use}
The work utilizes open-source models and resources that are in the public domain. The code, dataset, and associated models are released under the CC-BY Creative Commons license, in a GitHub repository that includes usage guidelines.

\subsection{Potential Risks}
We release a dataset from the 7th century. Many of the texts from that time period are biased, and some may find them offensive. The use of this dataset for metaphor detection does not appear to pose risks; however, it may result in biased or offensive models when it is used for other purposes.

\subsection{Annotator Guidelines} 
\label{appendix:annotator_guidelines}
\begin{enumerate}
    \item Metaphor could consist of one word or more.
    \item Metaphor cannot extend beyond the limits of a single poetic line. 
    \item An effort should be made to differentiate between different types of a metaphor, namely metonymy, synecdoche or personification. The top level of metaphor should be used if the distinction cannot be determined. \footnote{In this study, we have only used the metaphor/non-metaphor distinction,  but future versions will include metaphor subtypes.}

    \item Personifications of God should not be annotated as a metaphor unless the underlying personification is extended beyond its Biblical origin. 
    \item Payytanic epithets should be annotated only if they are based on a metaphor. That is to say, If the epithet is based solely on a paraphrase it is not metaphoric. 
\end{enumerate}

\begin{table*}[t]
    \centering
    \begin{tabular}{rll}
    \toprule
    Hebrew Source & Literal Translation & Meaning \\ \midrule 
    \<t.hd/s\{> \underline{\<\}.hwpt .hdryK>} & Room canopy will be renewed & Renovate the temple\\ 
    \underline{\<nhr lh.t>} \<mqwM>  & River place glow & River of fire \\
    \underline{\<w.tbyltm lh.t>} & Fiery immerse & Immerse yourself in a river of fiery fire\\

    \underline{\<h`yn>} \<wmr'yt> \underline{\<lb>} \<hrhwry> & Heart ponders eye sees & Sees and ponders \\ 
    \bottomrule
    \end{tabular}
    \caption{Examples from our dataset of sentences with expert disagreement in metaphor annotations. Metaphor labels are \underline{underlined} for annotator A and marked with \{brackets\} for annotator B.}
    \label{tab:disagreement}
\end{table*}

\subsection{Inter Annotator Disagreement} 
\label{appendix:iaa}
In order to better understand the discord between annotators, we will look at a few examples and discuss them. Examples of sentences with expert annotator disagreement appear in Table \ref{tab:disagreement}.
Looking at the first sentence in the table, the first annotator labeled only the first couplet as a metaphor and the verb at the end as a non-metaphor. According to the second annotator, the entire column is a metaphor (including the verb at the end). Though the central metaphor is the first two words, it is possible to interpret that the verb at the end, which refers to the metaphorical pronoun, also becomes metaphorical due to the context, but it is impossible to decide definitively.
With regards to the two next sentences, only one annotator thinks the phrase 'river of fire' is metaphorical while the other sees it as something literal. While in reality there is no river of fire, in the mythological view of the poet it is certainly something that can exist. Correspondingly, the question of whether it is possible to immerse in fire (as one immerses in water) cannot be given an unequivocal answer because if there is a river made of fire then surely one can immerse in it. As for the last row of the table, one annotator believes that the 'heart' and 'eye' represent the individual as a whole. In other words, it is not the heart that ponders nor the eye that sees, but the person who ponders and sees. Meanwhile, it is certainly possible to refer to them only in their simple sense and therefore not view them as metaphorical.

\subsection{Automatic Labeling}
\label{appendix:automatic_labeling}
For automatic labeling, we follow a BIO scheme, as common in other sequence labeling tasks like named entity recognition. In particular, the first word in each metaphor phrase is assigned a B-Metaphor tag, all other words in the same metaphor are assigned I-Metaphor, and all non-metaphor words are assigned O. This scheme allows us to perform word-level tagging and then convert back to multiword expressions, such that we can distinguish cases of two separate metaphoric words from a sequence of two words that constitute a metaphor.

When using Transformer models like AlephBERT or BEREL, words split into sub-word units, which are sequences of characters that do not necessarily correspond to meaningful morphemes. This is a data-driven splitting that is common in Transformer models, and we follow the same splitting as in the respective models (AlephBERT and BEREL).
The sub-word splitting has implications for training and testing the models. When training, we also follow a BIO scheme.
In particular, the first sub-word of each B-Metaphor is assigned a B-Metaphor tag, and all other sub-words in the same metaphor are assigned I-Metaphor. In the case of I-Metaphor and non-metaphor words, all sub-words are assigned with the original word tag.
At inference time, we predict tags for all sub-words, and if one sub-word received a metaphor tag (B-Metaphor or I-Metaphor), we determine that the word is a metaphor. We do this to prefer recall.

\subsection{Training Details} 
\label{sec:training}
In this study, there were two kinds of training: fine-tuning and model adaptation. Using transformers hyperparameter search, we found the best hyperparameters for fine-tuning. Refer to Table \ref{tab:hyperparameters} 
for the complete list of hyperparameters. We completed the hyperparameter search for each model and dataset pair. Since the hyperparameters were similar across experiments, we used the same hyperparameter throughout.
We repeated the experiments five times with seeds 41-45. The final results can be found in tables 
\ref{tab:results-pre}, \ref{tab:results-pinchas}.
The training was composed on Nvidia RTX 2080. A total of 16 experiments were conducted, five times each (different seeds), resulting in 13.5 hours of GPU time. 

For model adaptation, we used a learning rate of 1e-4, batch size 128, 3 epochs, and 10000 warmup steps.
The training was composed on Nvidia RTX 2080, with 10 hours of GPU time.

\begin{table}[h]
    \centering
    \begin{tabular}{lcc}
    \toprule
    & \multicolumn{1}{c}{Range} & \multicolumn{1}{c}{Best}  \\ 
    \midrule
    learning 

rate & $1e-6:1e-3$ & $5.4e-4$\\ 
    epochs & $2:10$ & $8$\\
    batch size & $16,32,64,128$ & $32$ \\  
    metaphor weight & $1:20$ & $9$ \\
    \bottomrule
    \end{tabular}
    \caption{Hyperparamets searched (range) and chosen (best) for fine-tuning. The metaphor weight is the weight for weighted cross entropy.}
    \label{tab:hyperparameters}
\end{table}

\subsection{Model Mistakes}
\label{appendix:mistakes}
Here we investigate the most common mistakes made by BEREL and AlephBert.
The most common false negative words in both models are \<n`ylt> (lock) and \</s`r> (gate). In the entire training set, the word ``gate'' appeared only five times as a metaphor (out of 20 times it appeared in the set). In the validation set, ``gate'' appears 27 times, 25 as a metaphor. The word ``lock'' did not appear at all in the training set, whereas it appears 25 times in the validation set, all of which were metaphorical. Interestingly, every time the word ``lock'' appears, it appears adjacent to the word ``gate''. While the AlephBERT model was wrong in 90 percent of the cases,  BEREL was wrong in 63 percent of the cases and predicted at least one word of the phrase as a metaphor in 72 percent of the cases.

The most common false positive among AlephBERT predictions is \<ybw'>  (come). The word appears 9 times in the training set, 2 of them metaphorically (22\%). In the validation set, it appears 29 times, all of them literally. AlephBert predicts that the use is literal 5 times correctly (17\%). BEREL, on the other hand, predicts correctly that the word is used literally in every case. The BEREL model was able to generalize better, probably since it learned important features for sentences in ancient Hebrew during the pre-training, whereas the AlepBERT model pre-trained model is less suitable for this language and probably learned some shortcuts, for example, the statistics of the word as a metaphor in the training set.

In Section \ref{sec:error_analysis} we provided an example of an error that could be attributed to inherent ambiguity rather than model inaccuracy.
Another example of this kind is \<l' ln.s.h q.sP> (not for eternity foam). The word foam also can be interpreted as 'angry', and thus can be interpreted as a metaphor or literal.
In both cases, these biblical metaphors are so common, that expert annotators refer to them as non-metaphors.

\subsection{Detailed Results} \label{sec:appendix-results}
Tables \ref{tab:results-pre} and \ref{tab:results-pinchas} show detailed results on both corpora, including accuracy, precision, and recall, in addition to F1 scores, which were given in the main body.

\balance

\begin{table*}[t]
    \centering
    \begin{tabular}{lcccc}
    \toprule
    Model & Accuracy & Precision & Recall & F1 \\ 
    \midrule
    Global majority & $82.5$ & $0.0$ & $0.0$ & $0.0$ \\ 
    Most frequent tag & $71.5$ & $48.5$ & $16.1$  & $24.2$ \\
    \midrule 
     BEREL random init & $78.5 \pm 1.3$ & $37.1 \pm 2.3$ & $26.6 \pm 4.0$ & $30.7 \pm 2.1$ \\    
    AlephBERT random init & $78.7 \pm 0.7$ & $37.3 \pm 1.4$ & $27.6 \pm 3.3$ & $31.6 \pm 2.2$ \\
    \midrule 
     BEREL & $\mathbf{82.2 \pm 0.4}$ & $\mathbf{51.1 \pm 1.4}$ & $38.2 \pm 1.2$ & $43.7 \pm 0.6$\\    
    BEREL WCE & $77.2 \pm 3.4$ & $41.7 \pm 3.9$ & $\mathbf{62.5 \pm 5.8}$ & $\mathbf{48.7 \pm 1.4}$ \\       
    AlephBERT & $78.5 \pm 2.0$ & $48.1 \pm 2.1$ & $35.6 \pm 3.7$ & $40.8 \pm 2.0$ \\  
    AlephBERT WCE & $76.2 \pm 0.1$ & $38.5 \pm 1.4$ & $56.4 \pm 2.6$ & $45.9 \pm 0.7$ \\
    AlephBERT adapted & $81.8 \pm 0.5$ & $49.4 \pm 2.0$ & $38.0 \pm 2.9$ & $42.8 \pm 1.3$\\
    AlephBERT adapted WCE & $76.2 \pm 1.7$ & $40.3 \pm 2.6$ & $59.5 \pm 4.4$ &  $47.2 \pm 0.9$  \\
    \bottomrule
    \end{tabular}
    \caption{Results on Pre-Classical Piyyut corpus: Average Accuracy, Recall, Precision, F1, and standard deviations for all described methods. Each experiment was repeated five times with different seeds. WCE refers to weighted cross-entropy loss.}
    \label{tab:results-pre}
\end{table*}

\begin{table*}[t]
    \centering
    \begin{tabular}{lcccc}
    \toprule
    Model & Accuracy & Precision & Recall & F1 \\ 
    \midrule
    Global majority & $79.7$ & $0.0$ & $0.0$ & $0.0$ \\ 
    Most frequent tag & $79.6$ & $49.9$ & $16.4$  & $24.7$ \\
    \midrule 
     BEREL random init & $73.2 \pm 2.6$ & $36.7 \pm 2.9$ & $33.1 \pm 6.3$ & $34.4 \pm 2.3$ \\    
    AlephBERT random init & $74.8 \pm 1.1$ & $36.5 \pm 1.7$ & $25.8 \pm 4.1$ & $31.3 \pm 3.4$ \\
    \midrule 
     BEREL & $\mathbf{79.7 \pm 1.1}$ & $\mathbf{53.6 \pm 4.1}$ & $41.6 \pm 5.2$ & $46.5 \pm 2.0$\\    
   BEREL WCE & $71.2 \pm 3.5$ & $40.0 \pm 2.9$ & $\mathbf{65.7 \pm 7.6}$ & $\mathbf{49.4 \pm 0.8}$ \\        
    AlephBERT & $79.1 \pm 0.8$ & $50.9 \pm 2.8$ & $36.1 \pm 1.7$ & $42.2 \pm 1.2$ \\ 
    AlephBERT WCE & $75.6 \pm 2.5$ & $43.9 \pm 3.9$ & $48.7 \pm 8.7$ & $45.5 \pm 2.0$ \\
    AlephBERT adapted & $79.7 \pm 0.9$ & $52.5 \pm 2.9$ & $39.3 \pm 2.4$ & $44.8 \pm 0.7$\\
    AlephBERT adapted WCE & $75.4 \pm 2.5$ & $43.9 \pm 3.6$ & $52.4 \pm 6.5$ &  $47.3 \pm 1.0$  \\
    \bottomrule
    \end{tabular}
    \caption{Results on Pinchas corpus: Average Accuracy, Recall, Precision, F1, and standard deviations for all described methods. Each experiment was repeated five times with different seeds. WCE refers to weighted cross-entropy loss.}
    \label{tab:results-pinchas}
\end{table*}

\end{document}